\documentclass[10pt,twocolumn,letterpaper]{article}

\usepackage[pagenumbers]{cvpr}

\usepackage{graphicx}
\usepackage{amsmath}
\usepackage{amssymb}
\usepackage{booktabs}
\usepackage{comment}
\usepackage{multirow}
\usepackage{tabularx}
\usepackage{float}
\usepackage{xcolor}
\usepackage{bbm}
\usepackage{import}
\usepackage{comment}
\usepackage{mathrsfs}

\makeatletter
\@namedef{ver@everyshi.sty}{}
\makeatother
\usepackage{tikz}

\usepackage{pifont} 
\usepackage{booktabs}
\usepackage{multirow}
\usepackage{adjustbox}
\usepackage{fontawesome} 
\usepackage{amsmath} 
\usepackage{url}

\def\link#1{
    \ifx&#1&
        \xmark{}
    \else
        {\href{#1}{\faExternalLink}}
    \fi
}

\renewcommand{\paragraph}[1]{\vspace{1em}\noindent\textbf{#1}.}
\newcommand{\exponential}[1]{\text{exp}\left(#1\right)}

\newcommand{\C}{\boldsymbol{C}}
\newcommand{\radiance}{\mathbf{c}}
\newcommand{\weight}{\alpha}
\newcommand{\density}{\sigma}
\newcommand{\trans}{\mathcal{T}}

\newcommand{\runtime}{10\xspace}

\newcommand{\methodtitle}{Instant Volumetric Head Avatars\xspace}
\newcommand{\model}{INSTA\xspace}

\usepackage[accsupp]{axessibility}
\usepackage[pagebackref,breaklinks,colorlinks]{hyperref}
\usepackage[capitalize]{cleveref}

\begin{document}

\title{\methodtitle}

\author{
Wojciech Zielonka
\qquad 
Timo Bolkart 
\qquad 
Justus Thies \\
Max Planck Institute for Intelligent Systems, Tübingen, Germany \\
{\tt\small \{wojciech.zielonka, timo.bolkart, justus.thies\}@tuebingen.mpg.de} 
}

\twocolumn[{%
\renewcommand\twocolumn[1][]{#1}%
\maketitle
\begin{center}
    \centering
    \captionsetup{type=figure}
     \includegraphics[width=\textwidth]{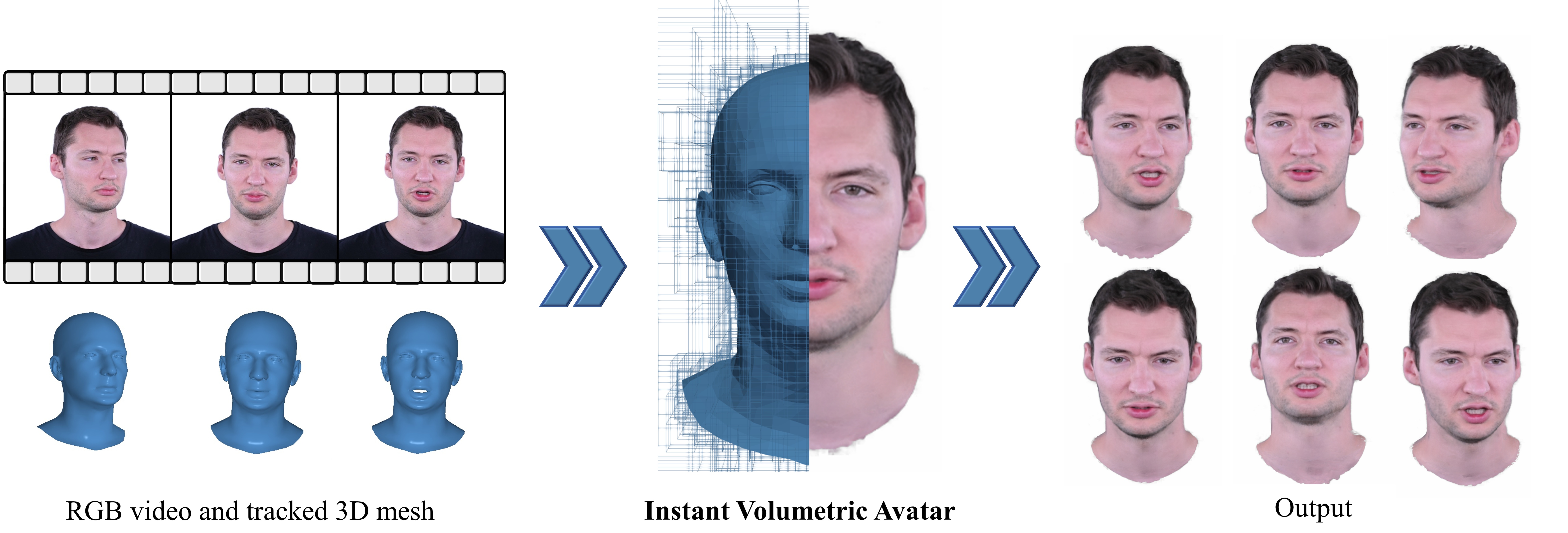}
      \caption{
      Given a short monocular RGB video, our method instantaneously optimizes a deformable neural radiance field to synthesize a photo-realistic animatable 3D neural head avatar.
      The neural radiance field is embedded in a multi-resolution grid around a 3D face model which guides the deformations.
      The resulting head avatar can be viewed under novel views and animated at interactive frame rates.
      }
      \label{fig:teaser}
\end{center}%
}]

\begin{abstract}
We present \methodtitle (\model), a novel approach for reconstructing photo-realistic digital avatars instantaneously.
\model models a dynamic neural radiance field based on neural graphics primitives embedded around a parametric face model.
Our pipeline is trained on a single monocular RGB portrait video that observes the subject under different expressions and views.
While state-of-the-art methods take up to several days to train an avatar, our method can reconstruct a digital avatar in less than \runtime minutes on modern GPU hardware, which is orders of magnitude faster than previous solutions. In addition, it allows for the interactive rendering of novel poses and expressions.
By leveraging the geometry prior of the underlying parametric face model, we demonstrate that \model extrapolates to unseen poses.
In quantitative and qualitative studies on various subjects, \model outperforms state-of-the-art methods regarding rendering quality and training time.
Project website: \href{https://zielon.github.io/insta/}{https://zielon.github.io/insta/}
\end{abstract}

\section{Introduction}
\label{sec:intro}
For immersive telepresence in AR or VR, we aim for digital humans (avatars) that mimic the motions and facial expressions of the actual subjects participating in a meeting.
Besides the motion, these avatars should reflect the human's shape and appearance.
Instead of prerecorded, old avatars, we aim to instantaneously reconstruct the subject's look to capture the actual appearance during a meeting.
To this end, we propose \methodtitle (\model), which enables the reconstruction of an avatar within a few minutes ($\sim$\runtime min) and can be driven at interactive frame rates.
For easy accessibility, we rely on commodity hardware to train and capture the avatar.
Specifically, we use a single RGB camera to record the input video.
State-of-the-art methods that use similar input data to reconstruct a human avatar require a relatively long time to train, ranging from around one day \cite{nha} to almost a week \cite{imavatar, nerface}.
Our approach uses dynamic neural radiance fields~\cite{nerface} based on neural graphics primitives~\cite{ngp}, which are embedded around a parametric face model~\cite{flame}, allowing low training times and fast evaluation.
In contrast to existing methods, we use a metrical face reconstruction~\cite{mica} to ensure that the avatar has metrical dimensions such that it can be viewed in an AR/VR scenario where objects of known size are present.
We employ a canonical space where the dynamic neural radiance field is constructed.
Leveraging the motion estimation employing the parametric face model FLAME \cite{flame}, we establish a deformation field around the surface using a bounding volume hierarchy (BVH)~\cite{bvh}.
Using this deformation field, we map points from the deformed space into the canonical space, where we evaluate the neural radiance field.
As the surface deformation of the FLAME model does not include details like wrinkles or the mouth interior, we condition the neural radiance field by the facial expression parameters.
To improve the extrapolation to novel views, we further leverage the FLAME-based face reconstruction to provide a geometric prior in terms of rendered depth maps during training of the NeRF~\cite{nerf}.
In comparison to state-of-the-art methods like NeRFace \cite{nerface}, IMAvatar \cite{imavatar}, or Neural Head Avatars (NHA) \cite{nha}, our method achieves a higher rendering quality while being significantly faster to train and evaluate.
We quantify this improvement in a series of experiments, including an ablation study on our method.

\medskip
\noindent
In summary, we present \methodtitle with the following contributions:
\begin{itemize}
    \item a surface-embedded dynamic neural radiance field based on neural graphics primitives, which allows us to reconstruct metrical avatars in a few minutes instead of hours or days,
    \item and a 3DMM-driven geometry regularization of the dynamic density field to improve pose extrapolation, an important aspect of AR/VR applications.
\end{itemize}

\section{Related Work}
\label{sec:related}
\begin{figure*}[ht!]
    \vspace{0.1cm}
    \includegraphics[width=\textwidth]{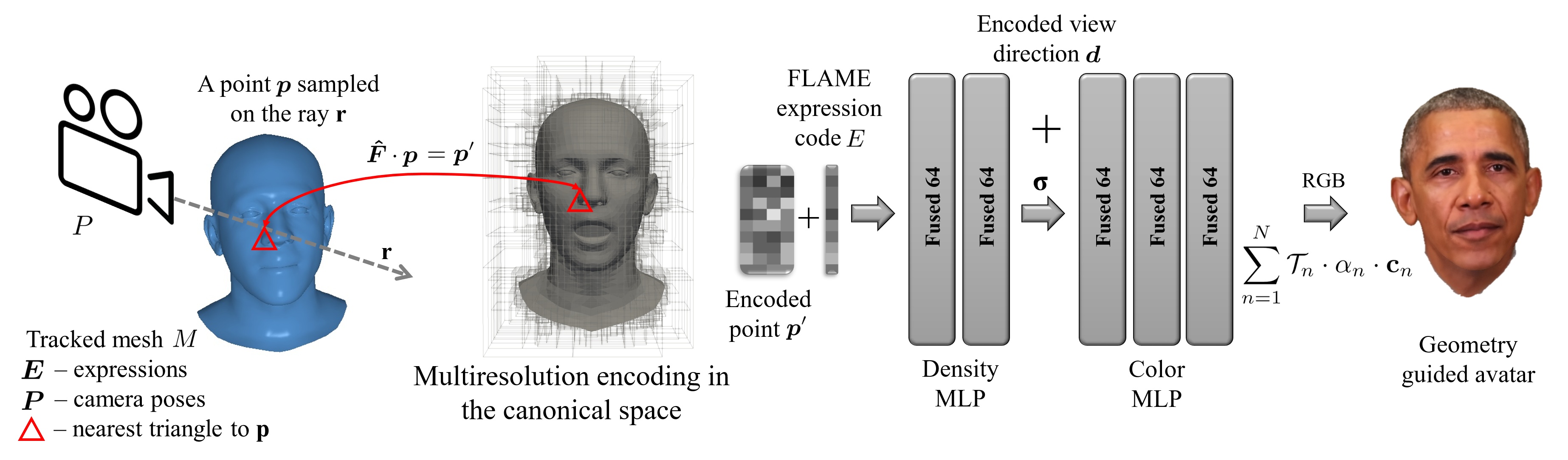}
    \caption{
    \textbf{Overview}. \model follows differentiable volumetric optimization introduced in \cite{nerf, ngp}.
    For each sampled point $\boldsymbol{p} \in \mathbb{R}^4$ in deformed space (in homogeneous coordinates), we are computing the nearest neighbor triangle on the mesh ${T_{def}} \in \boldsymbol{M}_i$ and its topological corresponding twin in the canonical space ${T_{canon}} \in \boldsymbol{M}^{canon}$.
    The deformation gradient of the triangle from deformed space to canonical space $\boldsymbol{\hat{F}} \in \mathbb{R}^{4\times4}$ defines the deformation field.
    Specifically, $\boldsymbol{p}$ is transformed to the canonical space by $\boldsymbol{p}' = \boldsymbol{\hat{F}}\cdot \boldsymbol{p}$.
    After canonicalization, the point is encoded using a multi-resolution hashing~\cite{ngp}.
    This feature is passed to fully fused multi-layer perceptrons  \cite{tiny-cuda-nn} with additional conditioning on the facial expressions $E_i$ and the encoded view direction $\boldsymbol{d}$.}

    \label{fig:pipeline}
\end{figure*}
\model is reconstructing animatable digital human avatars from monocular video data based on 3D neural rendering~\cite{tewari2022advances}.
Current solutions are using implicit representations \cite{nerf, nerface, nerfies, hypernerf, neural-actor, cao22} optimized via differentiable volumetric rendering, or are based on explicit models \cite{nha, f2f, online-modeling, exchaning-faces} for instance, triangle or tetrahedral meshes using differentiable rasterization~\cite{softras, jakob2022mitsuba3, chen2019dibrender, opendr}.
For a concise overview of neural rendering methods and face reconstruction, we point the reader to the state-of-the-art reports by Zollhöfer et al.~\cite{zollhoefer2018facestar}, and Tewari et al.~\cite{tewari2022advances,tewari2020neuralrendering}.

\vspace{-0.25cm}
\paragraph{Static Neural Radiance Fields}
Mildenhall et al.~\cite{nerf} and its many follow-up works~\cite{nerf++, liu2020neural, mip-nerf-360, mip-nerf, nerf-dark, ref-nerf, reg-nerf, roessle2021, block-nerf}, synthesize novel views of a complex static scene using differentiable volumetric rendering.
Many methods suffer from a long training time (1-5 days).
To this end, different acceleration methods have been proposed to improve the training time.
Yu et al.~\cite{plenoxels} achieved $100\times$ speedup by using a sparse voxel grid storing density and spherical harmonics coefficients at each node.
The final color is the composition of tri-linearly interpolated values of each voxel intersecting with the ray.
TensorRF~\cite{tensorf} factorizes the 4D NeRF scene into multiple compact low-rank tensor components achieving high performance and compactness.
The coordinate-based MLP is replaced with a voxel grid of features, and the final color is its vector-matrix outer product.
Müller et al.~\cite{ngp} introduced a new computer graphics primitive in the form of tiny MLPs which benefit from a multi-resolution hashing encoding.
The key idea is similar to Yu et al. \cite{plenoxels}.
The space is divided into an independent multi-level grid with feature vectors at the vertices of the grid.
A spatial hash function \cite{hash} is used to store the voxel grid efficiently.
Each point sampled on the ray is encoded by the interpolated feature vector of the corresponding grid level and passed to a tiny neural network to synthesize the final color.
Our method uses this efficient architecture to model the face in a canonical space.
Some of the static NeRF methods \cite{nerfing-mvs, depth-nerf, roessle2021,azinovic2021rgbdnerf} use additional depth maps to improve alignment and quality for static scenes. The depth priors help guide the ray sampling and better estimate the transmittance, resulting in improved geometry and color recovery.
While we are working with RGB images only, our method leverages the geometry prior of the 3DMM to guide the depth estimation during training, which results in an improved extrapolation ability w.r.t. view changes.

\vspace{-0.25cm}
\paragraph{Deformable Neural Radiance Fields}
After the introduction of NeRF~\cite{nerf} for static scenes, a natural research direction was to generalize it to dynamic, time-varying ones~\cite{li2021neural, du2021neural, dnerf, nerfies, hypernerf, nr-nerf}.
The reconstruction problem is divided into two different spaces, the deformed scene, and the canonical space, with a neural network as the mapper between them.
For human body modeling, a series of approaches have been proposed that leverage the kinematic chain of the SMPL~\cite{smpl} body model to condition the mapping function.
Peng et al.~\cite{animatable-nerf} proposed to learn blend weights to estimate the linear blend skinning-based warping field between canonical and deformed space based on the body skeleton.
Similarly, Neural Actor~\cite{neural-actor} uses a 3D body mesh proxy to learn pose-dependent geometric deformation and view-dependent appearance effects defined in the canonical space.
Lombardi et al.~\cite{volumetric-primitives}, which defines surface-aligned neural volumes to improve the rendering speed.
Garbin et al.~\cite{voltemorph} build a tetrahedral deformation graph around a radiance field based on the underlying mesh on which the deformations are defined, effectively transforming sampled points according to the current cage state.
Xu et al.~\cite{surface-aligned-nerf} propose surface-aligned neural radiance fields by projecting points in space to the surface of the body mesh. 
Our idea is based on a similar principle. However, instead of projecting points onto the mesh surface, we construct a 3D space around the head and deform points based on the deformation defined by the nearest triangles.
In contrast to modeling the deformation explicitly, Gafni et al.~\cite{nerface} implicitly model the facial expressions by conditioning the NeRF MLP with the global expression code obtained from 3DMM tracking~\cite{f2f} and by optimizing per latent frame codes to increase the network capacity for over-fitting.
In our approach, we leverage the idea of dynamic neural radiance fields to improve the mouth region's rendering, which is not represented by the face model motion prior.
Inspired by 3DMMs, IMAvatar \cite{imavatar} learns the subject-specific implicit representation of texture together with expression blendshapes and blend skinning weights.
They optimize an implicit surface by incorporating ray marching from Yariv et al. \cite{idr} with root-finding of the occupancy function \cite{snarf} to locate canonical correspondence of deformed points. However, we found the training time-consuming ($\sim$5 days) and unstable (can diverge).
In a concurrent work, Gao et al. \cite{gao2022reconstructing} create personalized blendshapes using neural graphics primitives, where for each of the blendshapes, a multi-resolution grid \cite{ngp} is trained.

\section{Instant Deformable Neural Radiance Field}
\label{sec:main}
Our goal is to create instant digital avatars which can be learned in a few minutes and rendered in interactive time.
For this purpose, we are using a geometry-guided deformable neural radiance field embedded into a multi-resolution hashing grid~\cite{ngp}, exploiting differentiable volumetric rendering~\cite{nerf} (see Fig.~\ref{fig:pipeline}).
For a given monocular video consisting of images $\boldsymbol{I} = \{I_i\}$ along with optimized intrinsic camera parameters $\boldsymbol{K} \in \mathbb{R}^{3\times3}$, tracked FLAME~\cite{flame} meshes $\boldsymbol{M} = \{M_i\}$ with corresponding facial expression coefficients $\boldsymbol{E} = \{E_i\}$ and poses $\boldsymbol{P} = \{P_i\}$, our goal is to build a controllable head avatar represented by a neural radiance field. 
To this end, we employ a canonical space where the neural radiance field is constructed.
To render specific facial expressions using volumetric rendering, we canonicalize the samples on a ray from the deformed space to query the radiance field in the canonical space.
%


%
\paragraph{Volumetric Rendering}
We take advantage of the recent advances in interactive NeRF optimization and use neural graphic primitives~\cite{ngp} to represent the radiance field.
The representation of the avatar is optimized using the differentiable volumetric rendering equation: 
\begin{align}
\hat{\C} = \int_0^D \trans(t) \cdot \density(t) \cdot \radiance(t) \; dt \;+\; \trans(D) \cdot \radiance_\text{bg},
\label{eq:volren}
\end{align}
where $\trans(t_n) = \exponential{- \int_{0}^{t_n} \density(t) \; dt} $ is the transmittance which indicates the probability of a ray traveling from $[0, t_n)$ without interaction with any other particles \cite{nerf}, $\density(t)$ is the density and $\radiance(t)$ is the radiance at position $\boldsymbol{p}_t$.
Note that the sample points $\boldsymbol{p}_t$ are canonicalized to access the actual radiance field.
Following NeRFace \cite{nerface}, we condition every sample $\boldsymbol{p}_t$ on the ray with the 3DMM facial expression code $E_i \in \mathbb{R}^{16}$ of video frame $i$.
Please note that in contrast to NeRFace \cite{nerface} and IMAvatar \cite{imavatar}, we do not use additional per-frame learnable codes.
The viewing vector $\boldsymbol{v} \in \mathbb{R}^{3}$ is encoded using spherical harmonics projection on four basis functions~\cite{axler2001harmonic, ngp} resulting in the final viewing vector encoding $\boldsymbol{d} \in \mathbb{R}^{16}$ which is concatenated with density MLP output.
While the viewing conditioning is applied on the entire avatar, the conditioning on facial expressions is bounded to the dynamically changing mouth region and is set to a constant vector $E_i = \textbf{1}$ for the other regions.
%

\paragraph{Canonicalization}
We define a mapping function $\Phi(\boldsymbol{p}, M_i)$ that projects a point $\boldsymbol{p} \in \mathbb{R}^4$ from the time-varying deformed space (where the volumetric rendering is performed) to the canonical space.
The mapping function leverages the time-varying surface approximation $M_i$ and a predefined mesh in canonical space $M^{canon}$.
We employ a nearest triangle search in deformed space to compute the deformation gradient $\boldsymbol{F} \in \mathbb{R}^{4\times4}$ which is used to map point $\boldsymbol{p}$ to the canonical counterpart $\boldsymbol{p'}$.
The deformation gradient $\boldsymbol{F}$ is computed via the known Frenet frames of the deformed triangle ${T_{def}}\in M_i$ and the canonical triangle ${T_{canon}} \in M^{canon}$.
Specifically, we compute the rotation matrices $\{{\boldsymbol{R}_{canon}}, {\boldsymbol{R}_{def}}\} \in \mathbb{R}^{3\times3}$ based on the corresponding tangent, bitangent, and normal vectors of a triangle.
With the translations $\{\boldsymbol{t}_{canon}, \boldsymbol{t}_{def}\} \in \mathbb{R}^{3}$ defined by a vertex of the triangle, they form the Frenet coordinate system frames
$\boldsymbol{L}_{canon}$ and $\boldsymbol{L}_{def} \in \mathbb{R}^{4\times4}$:
\begin{equation}
\begin{aligned}
    \boldsymbol{L}_{def} &= \begin{bmatrix}
      \boldsymbol{R}_{def} & \boldsymbol{t}_{def} \\ 
      \boldsymbol{0}^T & 1
    \end{bmatrix},\\
    \boldsymbol{L}_{canon} &= \begin{bmatrix}
      \boldsymbol{R}_{canon} & \boldsymbol{t}_{canon} \\ 
      \boldsymbol{0}^T & 1
    \end{bmatrix}.\\
\end{aligned}
\end{equation}
To account for any potential triangle size change between deformed and canonical spaces, we compute an isotropic scaling factor $\lambda \in \mathbb{R}$ via the relative surface area change of the given triangle w.r.t. its canonical twin $\lambda = \frac{a_{def}}{a_{canon}}$.
%
The deformation gradient $F$ is defined as:
\begin{equation}
\begin{aligned}
\boldsymbol{F} &= \boldsymbol{L}_{canon} \cdot \Lambda \cdot\boldsymbol{L}_{def} ^ {-1}, \\
\Lambda &= \begin{bmatrix}
  \lambda \boldsymbol{I} & \boldsymbol{0} \\ 
  \boldsymbol{0}^T & 1
\end{bmatrix}.
\label{eq:projection}
\end{aligned}
\end{equation}
%
%
%
To avoid transformation discontinuity, which arises from the local coordinate system of each triangle, we additionally perform exponentially weighted averaging of the transformations of the adjacent faces of the triangle's edges:
\begin{align}
    \boldsymbol{\hat{F}} = \frac{1}{\sum_{f \in A} \omega_f} \cdot \sum_{f \in A} \omega_f \boldsymbol{F}_f,
    \label{eq:projection_interp}
\end{align}
where $\omega_f = \exponential{-\beta || \boldsymbol{c}_f - \boldsymbol{p} ||_2}$, $\beta = 4$ and $A$ is the set of adjacent faces to $T$ (including $T$ with $\beta = 1$) with corresponding centroids $\boldsymbol{c}_f$. Please note that all vertex positions are defined in meters (FLAME metrical space).

To achieve interactive rendering as well as instantaneous optimization of the neural radiance field, we leverage a classical bounding volume hierarchy (BVH)~\cite{bvh} to significantly increase the nearest triangle search speed for the sampled points $\boldsymbol{p}_t$ on the ray. 
Note that methods like IMAvatar~\cite{imavatar} perform computation-heavy root-finding procedures to calculate surface points iteratively \cite{snarf}.
Our method builds a BVH based on the corresponding deformed mesh $M_i$ of frame $i$ to establish the mapping function to the canonical mesh.
Our BVH is implemented on GPU to utilize massively parallel nearest triangle search \cite{bvh-cuda}.
To alleviate the triangle search for highly tessellated FLAME regions, we simplified the eyeballs and the eye region~\cite{mesh-simplify}. Moreover, an additional set of triangles in the mouth region is used to serve as a deformation proxy (see sup. mat.).
%


\subsection{Training Objectives}

The optimization of the neural radiance field is based on a color reproduction objective and a geometry prior based on the 3DMM.
Following NeRF~\cite{nerf}, we redefine the volumetric rendering \Cref{eq:volren} with piece-wise constant density and color, and rewrite it in terms of alpha-compositing:
\begin{align}
\hat{\C}(t_{N+1}) = \sum_{n=1}^N \trans_n \cdot \weight_n \cdot \radiance_n,
\label{eq:final_color2}
\end{align}
where $\trans_n = \prod_{n=1}^{N-1}(1-\weight_n)$ weight $\weight_n$ is defined as $\weight_n \equiv 1-\exponential{-\density_n \delta_n}$ and $\delta_n$ is a step size equal $\frac{\sqrt{3}}{1024}$. To measure the photometric error, we use a Huber loss~\cite{huber-loss} with $\rho = 0.1$:
\begin{align}
\mathscr{L}_{color} =
    \left\{\begin{matrix}
        \frac{1}{2}(\C - \hat{\C})^{2} & if \left | (\C - \hat{\C}  \right | < \rho\\
        \rho ((\C - \hat{\C}) - \frac1 2 \rho) & otherwise
    \end{matrix}\right.
\end{align}
We enforce a depth loss to leverage the geometry prior of the reconstructed face based on the 3DMM FLAME.
Specifically, we rasterize the depth of the tracking mesh $M_i$ and apply an L1 distance between this map and the ray termination of the volumetric rendering.
As the FLAME model does not contain details like hair, we restrict the geometry prior to the face region:
\begin{align}
\mathscr{L}_{geom} = \sum_{\boldsymbol{r}} |\mathbbm{1}_{face}\{(z(\boldsymbol{r})-\hat{z}(\boldsymbol{r}))\}|,
\end{align}
where $\hat{z} = \sum_{n=1}^N \trans_n \cdot \weight_n \cdot t_{n}$, and $t_{n}$ is the current sample position, and $\mathbbm{1}_{face}\{\}$ is a segmentation indicator function which enables the loss for the face region.
The $\mathbbm{1}_{face}$ function uses face parsing information \cite{face-parsing} to decide a given pixel membership.
%
The total loss $\mathscr{L}$ is defined as:
\begin{align}
\mathscr{L} = \sum_{\boldsymbol{r}} \lambda_{color} (\boldsymbol{r})\mathscr{L}_{color}(\boldsymbol{r}) + \lambda_{geom}\mathscr{L}_{geom}(\boldsymbol{r}),
\label{eq:final_loss}
\end{align}
where $\lambda_{geom} = 1.25$ controls the influence of the geometry prior and $\lambda_{color} (\boldsymbol{r})$ weights the color loss contribution based on a face parsing mask.
Specifically, we weight the color loss higher for the mouth region with $\lambda_{color} = 40$ and $\lambda_{color} = 1$ otherwise. 

We implemented our animatable dynamic radiance field using the Nvidia NGP C++ framework~\cite{tiny-cuda-nn}.
We use two fully fused MLPs \cite{tiny-cuda-nn}, each with $64$ neurons, for color and density predictions.
The density MLP outputs feature values vector $\boldsymbol{\sigma} \in \mathbb{R}^{16}$ where the first value is the log-space density. The vector $\boldsymbol{\sigma}$ is later concatenated with the encoded viewing vector $\boldsymbol{d}$ to be the input of the color network.
For optimization, we used Adam \cite{adam} with an exponential moving average on the weights and fixed learning rate $\eta=2.5\mathrm{e}{-3}$.
In our experiments, we train the network for $32$k optimization steps.
We randomly sample 1700 frames from the whole dataset during the training and load them into the processing buffer.
Every 1500 steps, we repeat the procedure and resample the dataset.

\section{Dataset}
\label{sec:data}
\label{sec:data}
Our method takes a single video as input to generate the volumetric avatar of the depicted subject.
For our experiments, we recorded multiple actors with a Nikon Z6 II Camera as well as used sequences from Youtube, resulting in a set of twelve actors.
For the in-house recordings, we captured around 2-3min of monocular RGB Full HD videos, which later were cropped, sub-sampled to 25fps, and resized to ${512^2}$ resolution.
We additionally use background foreground segmentation using robust matting~\cite{robust-matting} and an off-the-shelf face parsing framework~\cite{face-parsing} for image segmentation and clothes removal.
\paragraph{Dataset Tracking Generation}
An essential part of this project is temporally stable face tracking of the monocular input data.
To this end, we use the analysis-by-synthesis-based face tracker from MICA~\cite{mica}, based on Face2Face~\cite{f2f} using a sampling-based differentiable rendering. We refer to the original paper \cite{f2f} for more details.
We extend the optimization with two extra blendshapes for eyelids and iris tracking using Mediapipe~\cite{mediapipe}.
In contrast to MICA, we also optimize for FLAME shape parameters, with regularization towards MICA shape prediction instead of the average face shape as in Face2Face~\cite{f2f}.
Note that for our prototype, we implemented the tracking in PyTorch, which is significantly slower than the original Face2Face implementation, which can track faces in real-time.

\section{Results}
\label{sec:results}
\begin{figure}[t!]
    \centering
    \includegraphics[width=\linewidth]{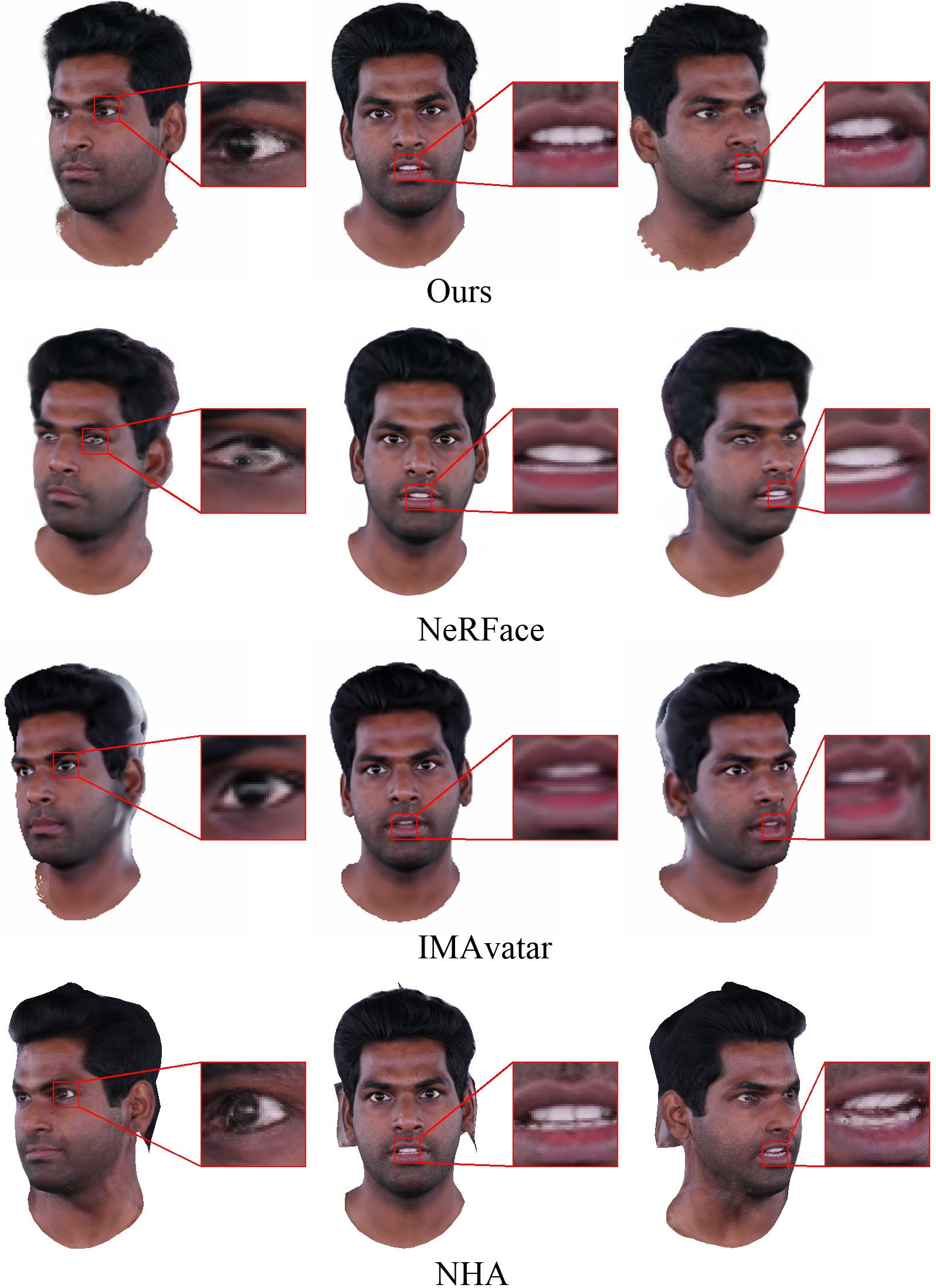}
    \caption{Qualitative comparison for novel view extrapolation. As can be seen, our method can better handle image synthesis under novel poses.
    NHA \cite{nha} suffers from degenerated geometry with many artifacts at the ear region. NeRFace \cite{nerface} lacks high-frequency details for eyes and teeth, and IMAvatar \cite{imavatar} shows silhouette artifacts at gracing angles.
    }
    \label{fig:novel_views}
\end{figure}

\begin{figure*}[ht!]
    \centering
    \vspace{-1.25cm}
    \includegraphics[width=\textwidth]{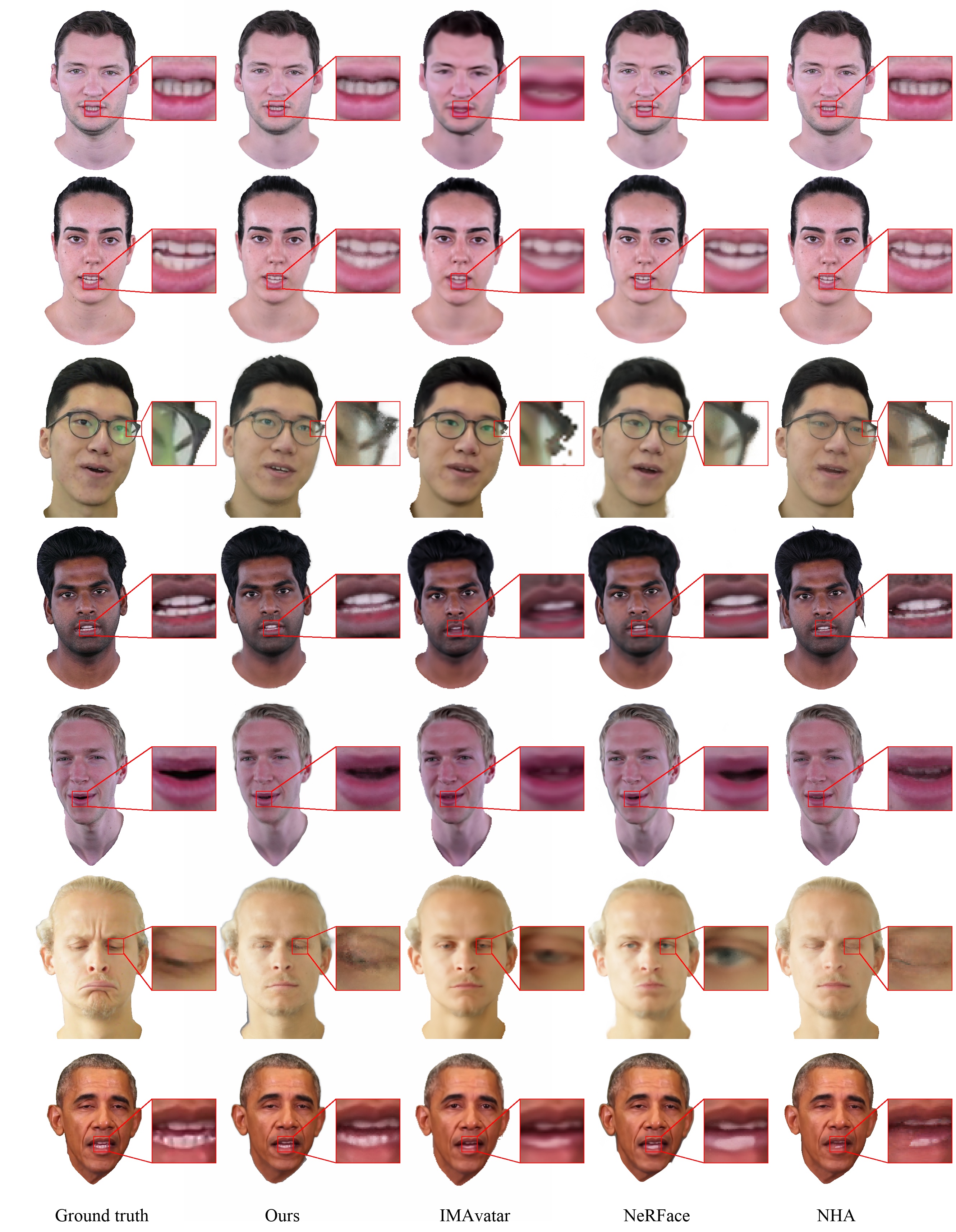}
    \vspace{-0.5cm}
    \caption{
    Qualitative comparisons show that our method produces high-quality facial avatars which beat the state-of-the-art methods in terms of image quality (e.g., capturing fine details like lips and teeth) while being significantly faster to obtain.
    }
    \label{fig:main_results}
\end{figure*}

In this section, we evaluate the quality of the synthesized digital human avatars generated by our method \model in comparison to state-of-the-art.
For this purpose, we use the test sequences from our dataset, which consist of the last 350 frames of each video.
%
%

\begin{figure*}[t!]
    \includegraphics[width=\textwidth]{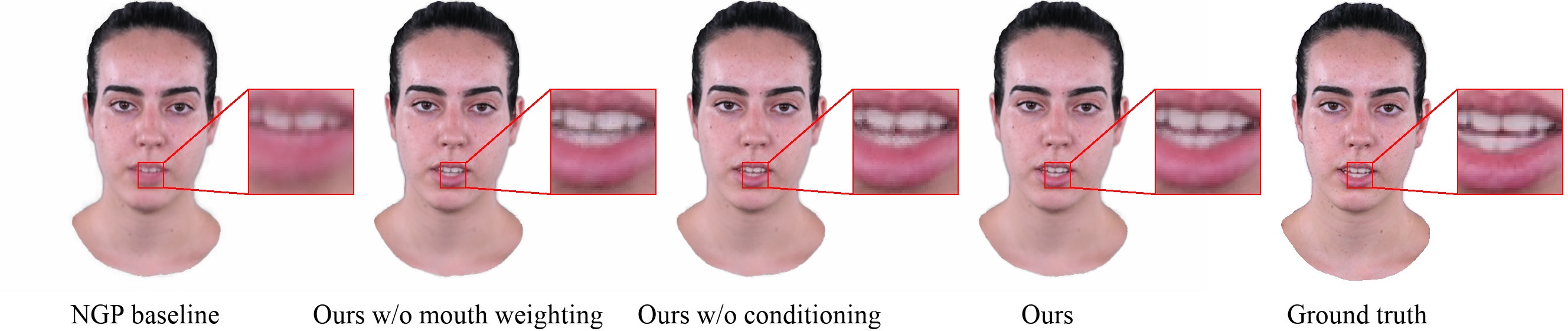}
    \caption{Embedding the neural radiance field around the deformable face model allows us to model dynamic sequences in contrast to the static radiance field of NGP \cite{ngp}.
    The expression conditioning and face-parsing-based weighting leads to sharper teeth reconstructions.
    }
    \label{fig:ablation_color}
\end{figure*}


\subsection{Image Quality Evaluation}
To evaluate our method in terms of the image quality and novel view extrapolation, we make a comparison to NeRFace \cite{nerface}, IMAvatar \cite{imavatar}, and Neural Head Avatars (NHA) \cite{nha}.
For this comparison, we use the original implementations of the authors.
Note that for IMAvatar, we use the most recent version of the author's code, which contains additional semantic information for mouth interior and FLAME geometry supervision which is different from the original paper.
\Cref{fig:main_results} depicts qualitative results evaluated on the test sequences.
To evaluate the image quality of the results quantitatively, we use several pixel-wise metrics; mean squared error, SSIM, PSNR, and the perceptual metric LPIPS~\cite{lpips} (see \Cref{tab:color_errors}).
Note that IMAvatar is trained at a resolution of $256^2$ due to its computational complexity; for the comparison, we upsample the results to $512^2$.
All methods produce sharp and photo-realistic images which are hard to distinguish from the ground truth.
However, the most noticeable artifacts, especially for the ear regions, were generated by NHA.
Moreover, IMAvatar, for some of the videos, had problems with convergence and stability, leading to diverging optimization and premature termination of the training.
Compared to these methods, our approach can achieve the best image quality while being significantly faster to train (see sup. mat.).
Extrapolation to novel views is an essential aspect of 3D digital avatars that are used in AR or VR applications.
In \Cref{fig:novel_views}, we depict a viewpoint extrapolation comparison with the baseline methods.
We can observe that NeRFace~\cite{nerface} produces blurry results in the area of eyes and teeth.
IMAvatar~\cite{imavatar} exhibits artifacts at gracing angles at the silhouette, and NHA~\cite{nha} suffers from degenerated geometry with strong artifacts at the ears.
In contrast to these methods, our method can robustly generate photo-realistic images under novel poses and achieves high visual quality, especially in the skin and mouth region.

\begin{table}[t!]
    \centering
    \resizebox{1.0\linewidth}{!}{
    \begin{tabular}{lccccr}
        Method & L2 $\downarrow$ & PSNR $\uparrow$ & SSIM $\uparrow$ & LPIPS $\downarrow$ & Time $\downarrow$ \\
        \midrule
        NHA \cite{nha} &   0.0022 &  27.71 &  \textbf{0.95} &  \textbf{0.04} & 0.63 \\
        IMAvatar \cite{imavatar} &  0.0023 &  27.62 &  0.94 &  0.06 & 12.34 \\
        NeRFace \cite{nerface} & \textbf{0.0018} &  \textbf{29.28} &  \textbf{0.95} &  0.07 & 9.68 \\
        Ours           &  \textbf{0.0018} &  28.97 &  \textbf{0.95} &  0.05 & $\textbf{0.05}$ \\
        \bottomrule
    \end{tabular}}
    \caption{
        Average photometric errors over 19 videos from our dataset, NHA, IMAvatar, and NeRFace datasets (see \cref{fig:main_results}). The average rendering time of a single frame in seconds is denoted as \emph{Time} in the rightmost column.
        Our method is on par with NeRFace of Gafni et al. w.r.t. the pixel-wise error metrics. Additionally, our approach achieves low perceptual error in comparison to all methods while being significantly faster to train and evaluate.
        }
    \label{tab:color_errors}
\end{table}


\subsection{Ablation Studies}

\begin{figure}[ht!]
    \centering
    \includegraphics[width=0.9\linewidth]{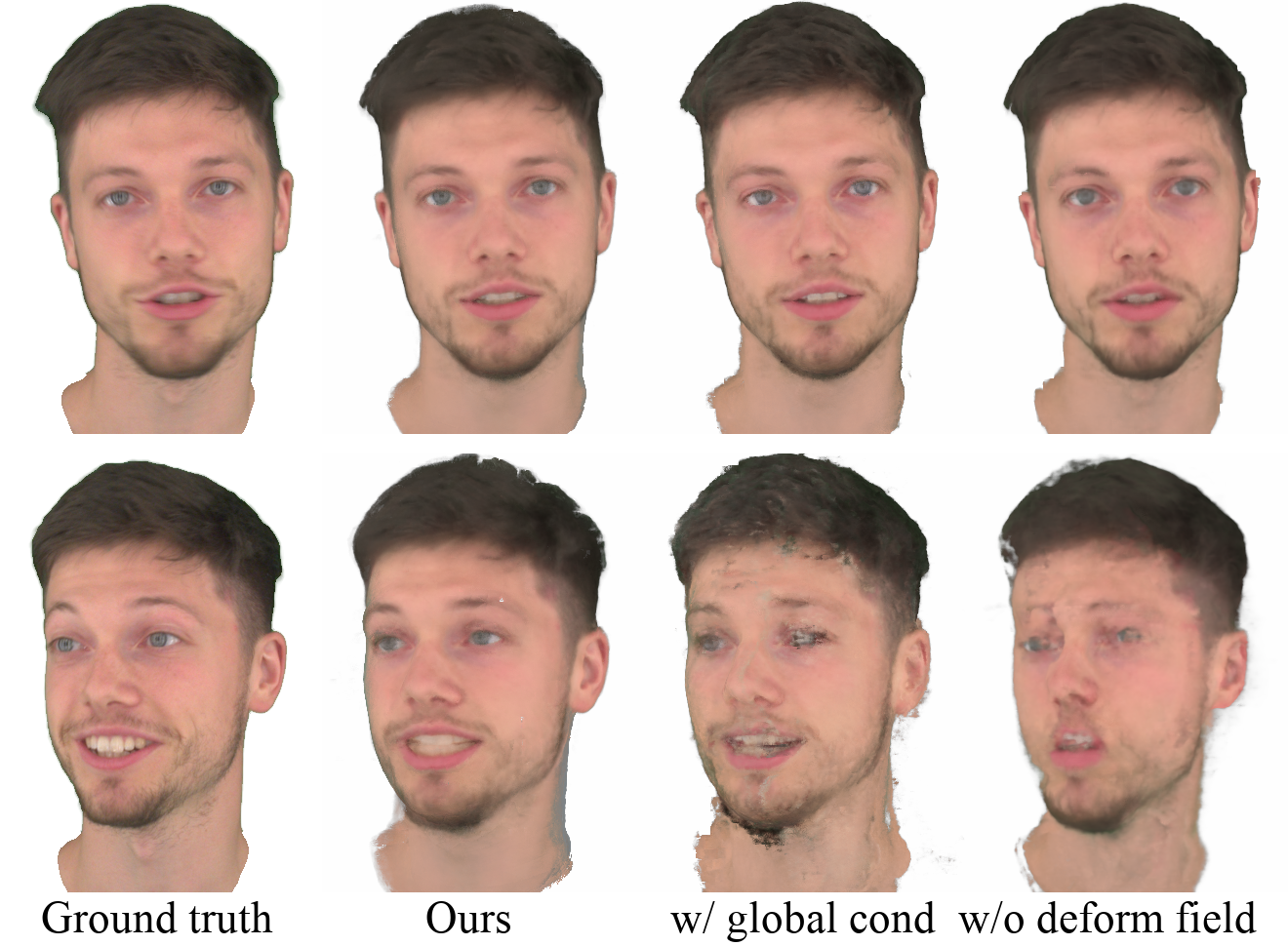}
    \caption{Ablation study w.r.t. the conditioning and deformation field. From left to right: ground truth, ours, ours with global conditioning, and ours without deformation field but with per-frame learnable codes (NeRFace).}
    \label{fig:ablation_conditioning}
\end{figure}

\begin{figure}[ht!]
    \centering
    \includegraphics[width=0.8\linewidth]{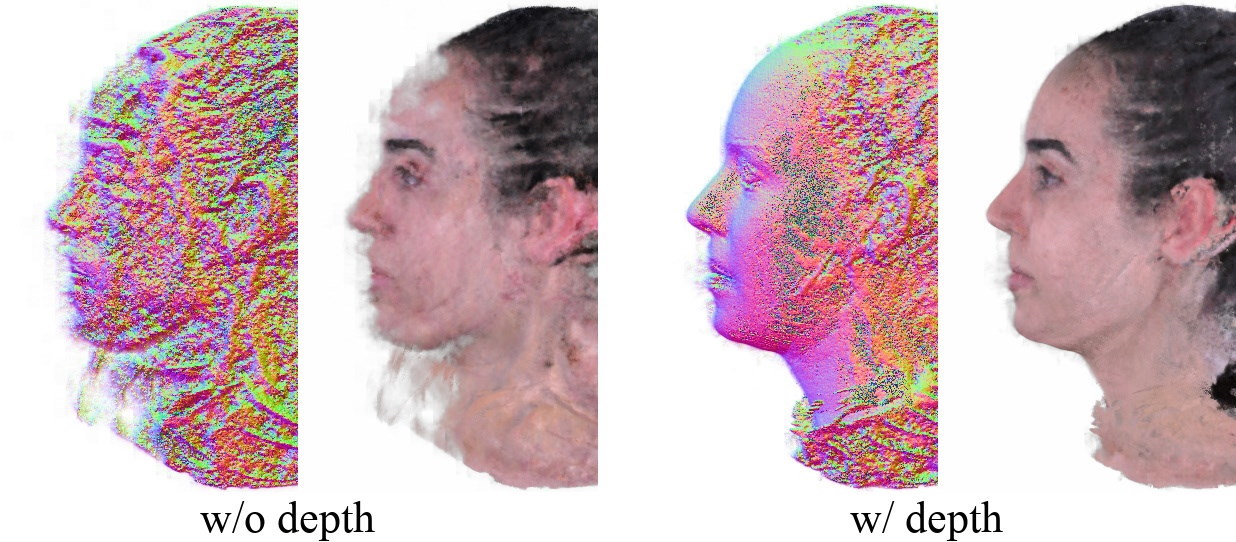}
    \caption{
    The geometric prior of the 3DMM helps for extrapolation to extreme novel views, in this case, $90^{\circ}$.
    }
    \label{fig:ablation_depth}
\end{figure}

We conducted a series of ablation studies to analyze the different components of our pipeline. 
Specifically, we are interested in the influence of localized expression conditioning for teeth quality (\Cref{fig:ablation_color}), the effect of the geometric prior (\Cref{fig:ablation_depth}), especially for the novel view synthesis, and the importance of the deformation field (\Cref{fig:ablation_conditioning}).

\paragraph{Deformation Field}
\Cref{fig:ablation_conditioning} shows the impact of the deformation field and the conditioning on the quality of the renderings.
We conducted two experiments where we used 
\textbf{a)} a global conditioning instead of the local one and \textbf{b)} global conditioning with per-frame learnable codes and without the deformation field (similar to NeRFace).
As can be seen, local conditioning and the mesh-based deformation field helps to avoid overfitting to the short training sequences.

\begin{figure*}[ht!]
    \centering
    \includegraphics[width=\linewidth]{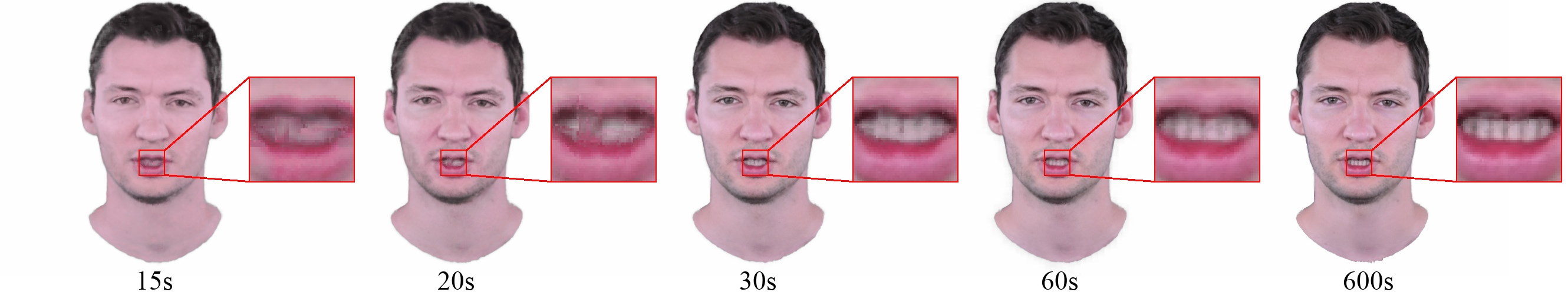}
    \caption{
    \model allows training personalized volumetric avatars from RGB videos within a couple of seconds.
    Already after 30 seconds of optimization, we achieve good results where the geometry and appearance match the input.
    To improve the reconstruction of high-frequency details like teeth, the method needs to train approximately \runtime min.
    }
    \label{fig:ablation_time}
\end{figure*}

\paragraph{Geometric Prior}
We leverage the geometric prior of the 3DMM FLAME~\cite{flame} to regularize the depth estimations of our volumetric rendering method.
During training, we render depth maps of the per-frame 3DMM reconstructions and measure a loss between the estimated ray termination and the depth of the rendered face model.
In \Cref{fig:ablation_depth}, we show an ablation study w.r.t. this geometric prior.
The generated digital avatar is shown from an unseen profile view, an extreme extrapolation from the training data which observed views in a range of $\pm40^\circ$.
Using the additional geometric prior improves the stability and quality of the results.

\paragraph{Expression Conditioning}
Most publicly available 3DMMs \cite{flame, bfm} do not explicitly model teeth.
However, this region is especially challenging for the reconstruction of 3D facial avatars due to highly dynamic lips, which can occlude the teeth depending on the given expressions.
To compensate for the missing geometry, we condition this region on FLAME expression coefficients.
In \Cref{fig:ablation_color}, we show that using this additional information helps to improve the synthesis of the mouth interior.
Furthermore, we demonstrate that a higher color term weight on the mouth region (\Cref{eq:final_loss}) improves the visual quality.

\section{Discussion}
\label{sec:discussion}
While our method \model shows better quality and speed compared to state-of-the-art RGB-video-based avatar generation techniques, there are still several challenges that need to be addressed in future work. 
Our model handles the dynamically changing facial expressions but does not capture dynamically changing hairs. Thus, the hair quality is not on par with the face interior and still needs improvements in the level of detail.
Furthermore, the used 3DMM does not model teeth geometry. A better approximation of the mouth region would increase the viewpoint extrapolation with improved quality of teeth.
While our method achieves real-time frame rates for rendering at a resolution of $512^2$, the rendering speed needs to be improved to enable high-quality video conferences in AR or VR, especially when a higher resolution is required.
With additional engineering, the training process of our method could be moved to a background process that would continuously refine our canonical avatar after an initial warm-up stage.
For example, regions initially not visible could be captured during the conversation, and the avatar would be updated accordingly.

\section{Limitations}
\label{sec:limitations}
\begin{figure}[ht!]
    \centering
    \includegraphics[width=\linewidth]{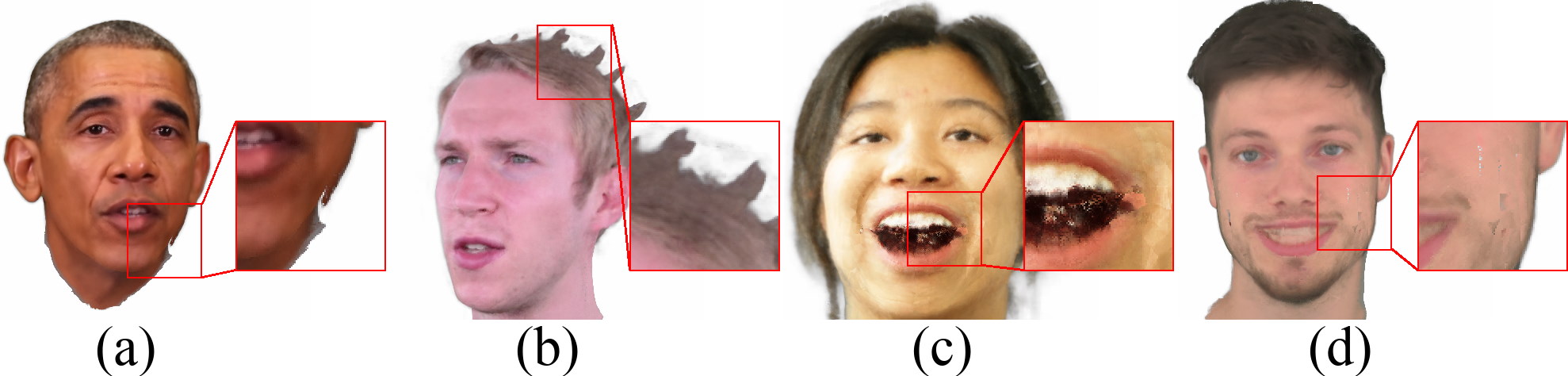}
    \caption{Failure cases: \textbf{(a)} and \textbf{(b)} exhibits outline artifacts at the chin and hair which stem from geometry misalignment of the tracker, \textbf{(c)} extreme expressions can cause artifacts in the mouth region, and \textbf{(d)} extrapolation of expressions can lead to artifacts.}
    \label{fig:failure}
\end{figure}

An important quality factor of our method is face tracking, as misalignments of the geometry and the images will be propagated to the final avatar.
Another limiting aspect is the mouth interior quality due to the lack of geometry in that region, as can be seen in \Cref{fig:failure}.

\section{Conclusion}
\label{sec:conclusion}
\methodtitle (\model) is a novel approach that instantaneously optimizes geometry-guided 3D digital avatars.
Our method takes a monocular RGB video as input and optimizes a subject's dynamic neural radiance field in less than \runtime minutes using neural graphics primitives embedded around a 3DMM.
In comparisons and ablation studies, we demonstrate the capabilities of \model, which enable us to instantaneously create avatars that reflect reality and not a prerecorded appearance that might deviate from the current look of the person.
We believe this paradigm change to adaptable online avatars is a stepping stone toward immersive telepresence applications.

\paragraph{Acknowledgement}
The authors thank all participants of the study and the International Max Planck Research School for Intelligent Systems (IMPRS-IS) for supporting WZ.
While TB is a part-time employee of Amazon, his research was performed solely at and exclusively funded by MPI.
JT is supported by Microsoft Research gift funds.

{\small
\bibliographystyle{splncs04}
\bibliography{main}
}

\clearpage
\appendix 
\twocolumn[{%
\renewcommand\twocolumn[1][]{#1}%
\begin{center}
     \textbf{\Large{Instant Volumetric Head Avatars \\ -- Supplemental Document --}}
     \vspace{1cm}
     \centering
     \captionsetup{type=figure}
     \includegraphics[width=0.9\textwidth]{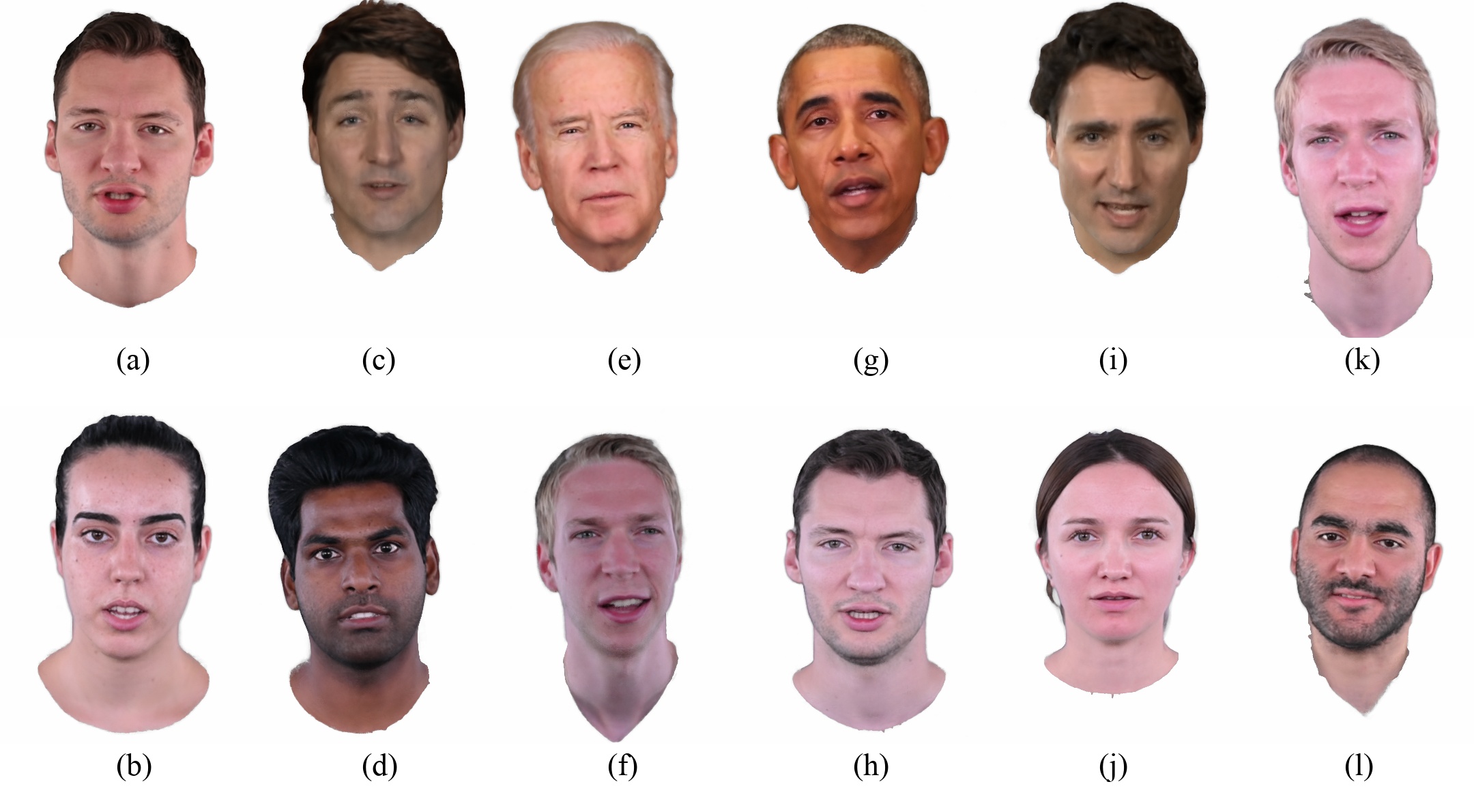}
     \caption{Our dataset consists of twelve sequences obtained from both in-house recordings and YouTube. Here, we present their respective volumetric avatars which \model optimizes in less than \runtime minutes. As can be seen, our method works well in both cases producing photorealistic videos even for in-the-wild sources (c, e, g, i). Please see the supplemental video for animations of these avatars.}
      \label{fig:supp-teaser}
\end{center}%
}]

This document elaborates on additional results and potential applications beyond volumetric video conferencing.
Specifically, we cover details about the acceleration structures for ray sampling with respect to the neural graphic primitives~\cite{ngp} (see \Cref{sec:accelerated}).
We demonstrate facial expression transfer in \Cref{sec:applications}.
In \Cref{sec:suppablation}, we show additional qualitative and quantitative results in terms of predicted normal and error maps.

\section{Implementation Details}
\label{sec:accelerated}

\paragraph{Accelerated Ray Marching}
\model is based on NGP~\cite{ngp}, which achieves significant speedup by adapting the sampling strategy utilizing occupancy grids.
For a given scene, a separate gird of $128^3$ is used to store an occupancy bit.
During ray marching for a given sample point, a bit value is measured to determine if this position should be skipped.
In this way, samples in empty spaces can be effectively omitted.
The occupancy grid is continually updated during the training based on the density values that can be predicted from the neural graphic primitives.
To accommodate dynamic scenes, we adapted this mechanism.
Specifically, we construct the acceleration structure in the deformed space where we shoot rays, which is different from the canonical space where the neural graphic primitives are learned.
Throughout the training, the acceleration structure converges to a Boolean union across all expressions in the training dataset.
Optionally, during the update step of the acceleration structure, the occupancy bit for voxels that lay on the isosurface around the canonical mesh can be manually set to on thus creating a fixed sampling region. This approach impedes the rendering time. However, sequences with very expressive motion can benefit from it, especially for motion extrapolation.
In \Cref{tab:hyperparameters}, we detail the hyperparameters of the hashing grid used to store the radiance field.

\begin{table}[h!]
    \centering
    \medskip
    \begin{tabular}{ll}
        Parameter &     Value \\
        \midrule
        Number of levels & 16  \\
        Hash table size & $2^{18}$  \\
        Number of features per entry & 8 \\
        Coarsest resolution & 16  \\
        Finest resolution & 2048 \\
        \bottomrule
    \end{tabular}
    \caption{
    Hyperparameters used for the hash encoding grid \cite{ngp}.}
    \vspace{-0.25cm}
    \label{tab:hyperparameters}
\end{table}

\begin{figure}[ht!]
    \centering
    \includegraphics[width=0.65\linewidth]{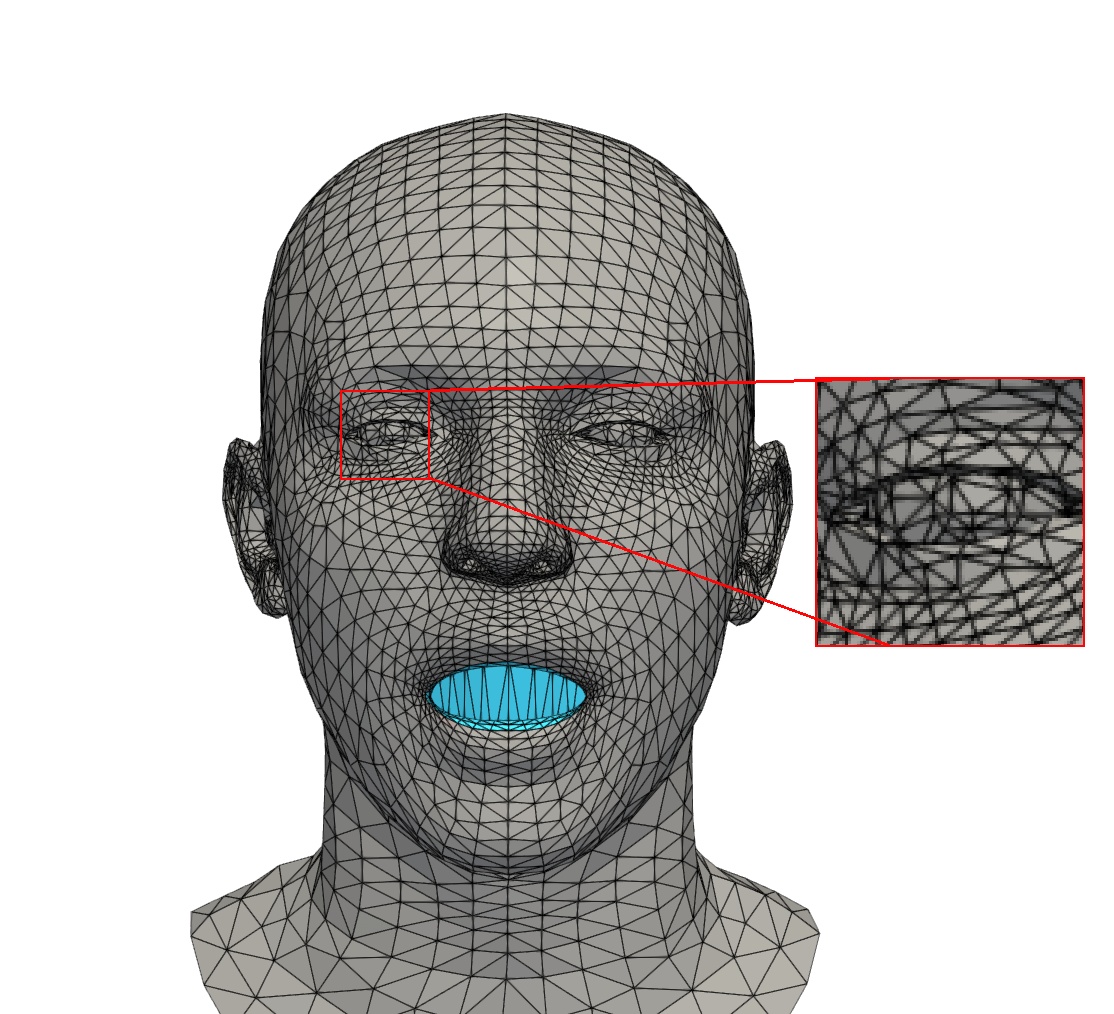}
    \vspace{-0.2cm}
    \caption{
    For the expression conditioning, we only consider the mouth region as depicted in magenta color. Moreover, we simplify the eyeballs, since they are modeled as densely tessellated spheres which introduce unnecessary computation overload for nearest neighbor search.
    }
    \vspace{-0.5cm}
    \label{fig:masks}
\end{figure}

\paragraph{FLAME masks}
Our method uses a mask defined on the simplified FLAME topology complemented with an additional set of triangles for the mouth cavity which is used for expression conditioning (Section 3, main paper).
\Cref{fig:masks} depicts the conditioning region which handles the dynamic appearance of the mouth interior. Additionally, the new eyeballs topology is magnified. For the mesh simplification, we used Garland et al. \cite{mesh-simplify} to compute a new set of faces which we later transferred to all meshed for a given sequence.


\section{Applications}
\label{sec:applications}

Our volumetric avatars have many potential applications.
Because they are controlled by a parametric face model, the expression transfer can be easily applied. We follow NeRFace \cite{nerface} and calculate relative expressions by manually selecting a neutral face of the target $T_{neutral}$ and source $S_{neutral}$ for which we calculate delta expressions $\Delta_{i} = S_{i} - S_{neutral}$. Finally, those relative delta vectors for each frame can be transferred to the target by $T_{i} = T_{neutral} + \Delta_{i}$. This step is necessary due to the fact that 3DMMs do not completely disentangle identity and expressions, thus transferring directly the expression coefficients from one actor to the other will change the mesh shape.

\begin{figure}[ht!]
    \centering
    \includegraphics[width=\linewidth]{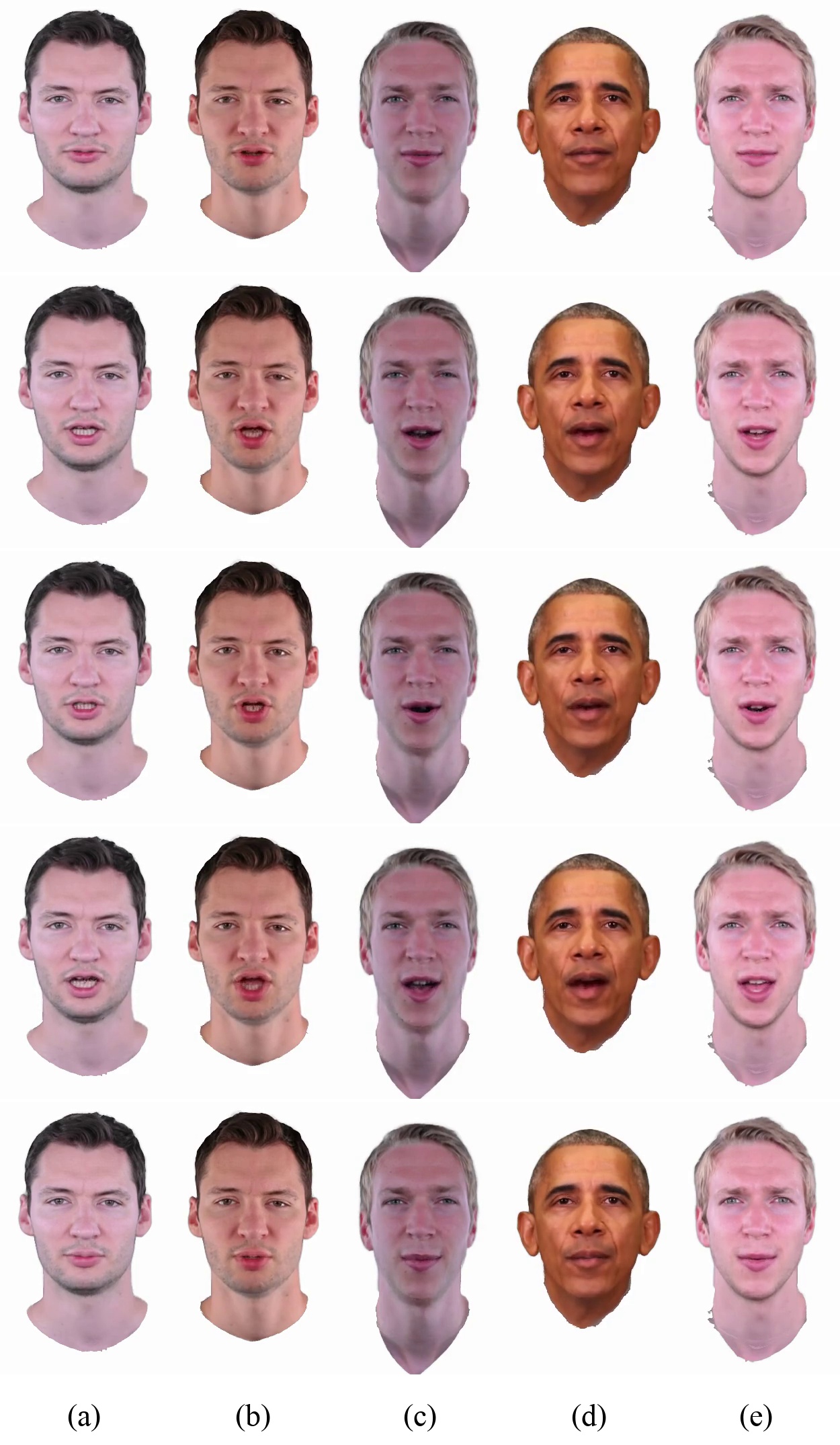}
    \caption{Expression transfer: (a) source actor, (b-e) target subjects with expression from (a).}
    \vspace{-0.5cm}
    \label{fig:retargeting}
\end{figure}

Expression transfer has been demonstrated in a variety of state-of-the-art methods on facial avatar reconstruction~\cite{f2f, nerface, real-time-exp-transfer, nha, imavatar}.
Specifically, the expression from a source subject is captured and then applied to a facial avatar of a different person.
\Cref{fig:retargeting} shows that our method can be used for such an application.

\section{Additional Results}
\label{sec:suppablation}

\begin{figure*}[ht!]
    \centering
    \includegraphics[width=0.95\textwidth]{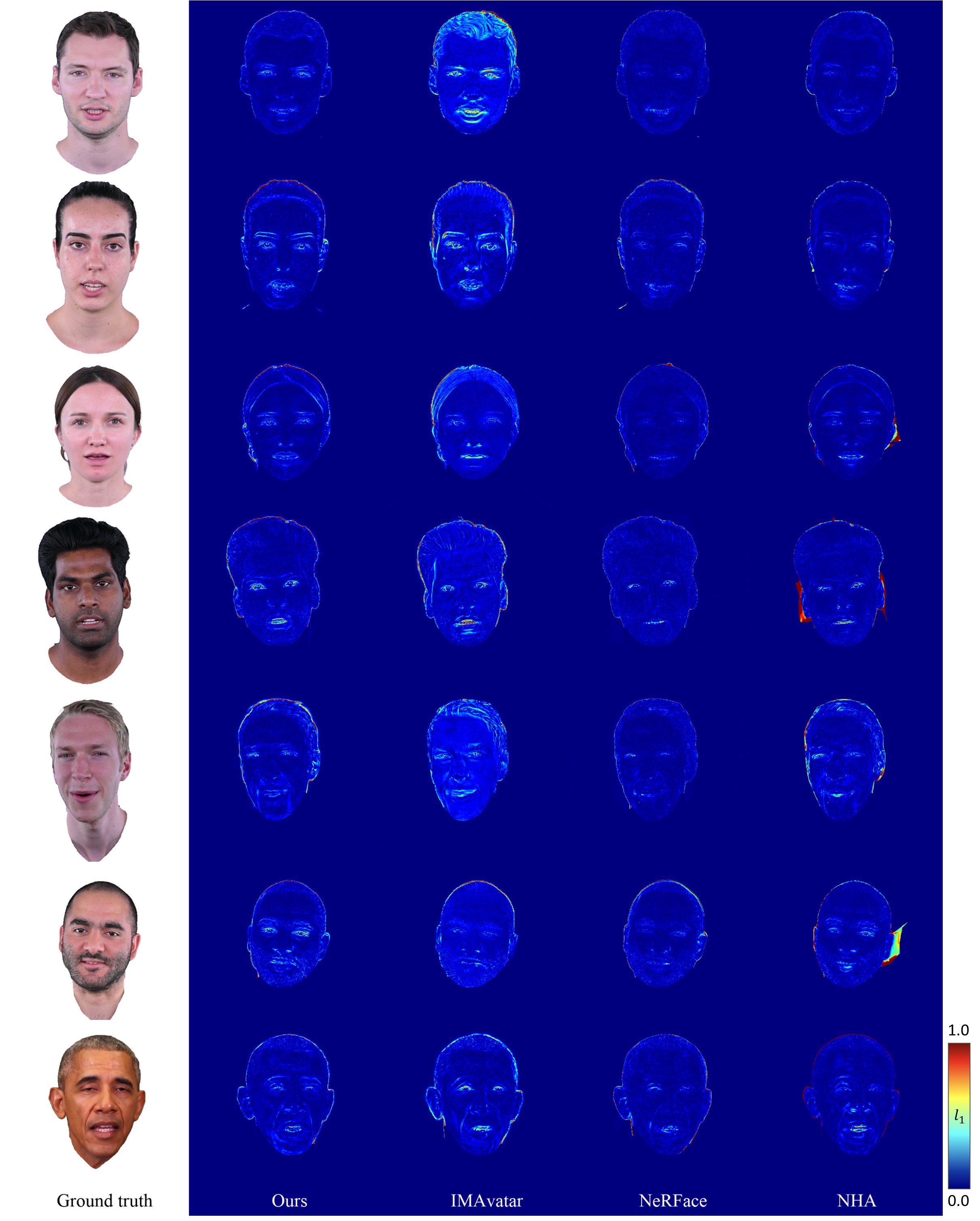}
    \caption{
    The heatmaps based on $l_1$ RGB distance represent photo metric errors on the test sequences.
    IMAvatar \cite{imavatar} synthesizes images with a low level of detail. Geometry mispredictions of NHA \cite{nha} create artifacts around the ear region.
    }
    \label{fig:heatmaps}
\end{figure*}

\begin{figure*}[ht!]
    \centering
    \subfloat[]{
        \includegraphics[width=0.45\linewidth]{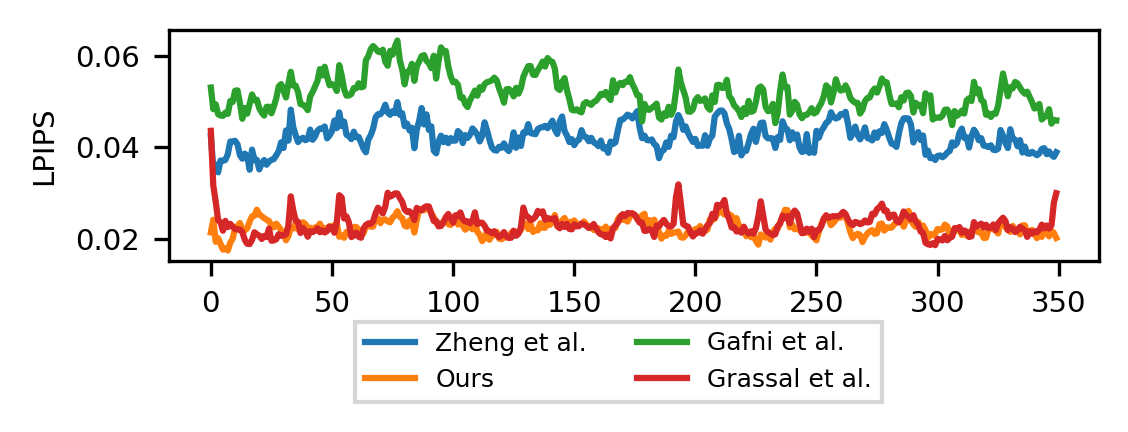}
    }
    \subfloat[]{
        \includegraphics[width=0.45\linewidth]{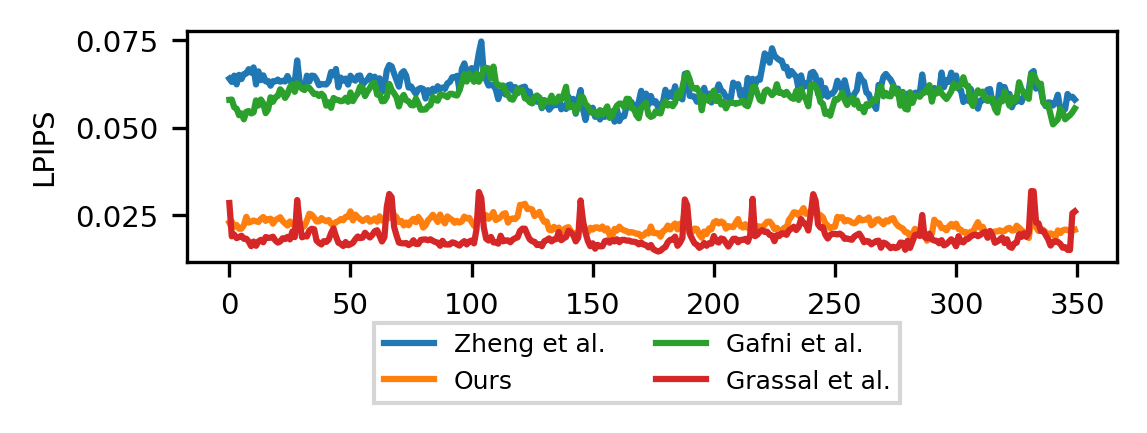}
    }
    \vfill
    \subfloat[]{
        \includegraphics[width=0.45\linewidth]{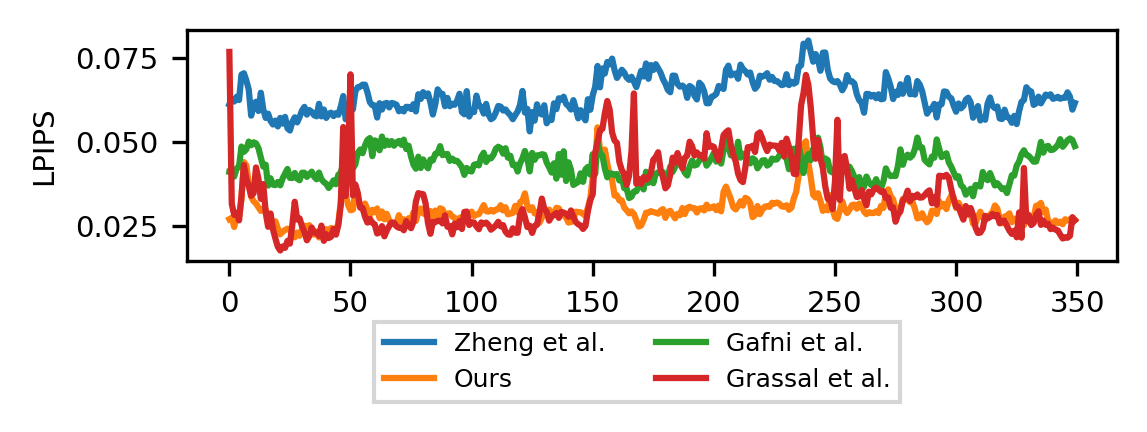}
    }
    \subfloat[]{
        \includegraphics[width=0.45\linewidth]{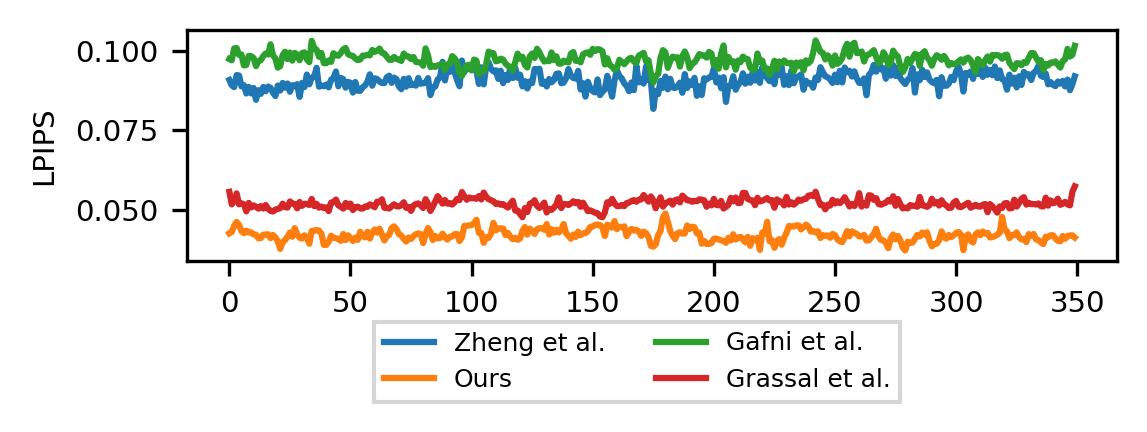}
    }
    \vfill
    \subfloat[]{
        \includegraphics[width=0.45\linewidth]{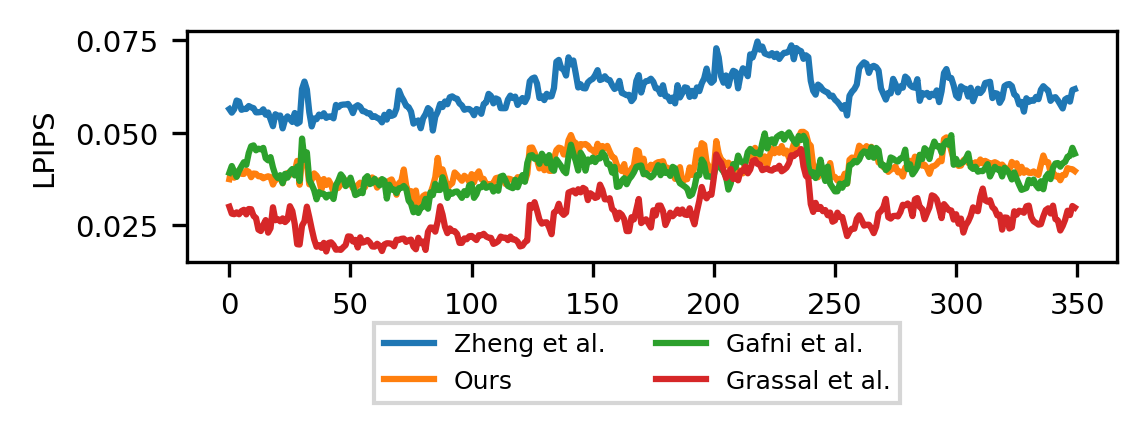}
    }
    \subfloat[]{
        \includegraphics[width=0.45\linewidth]{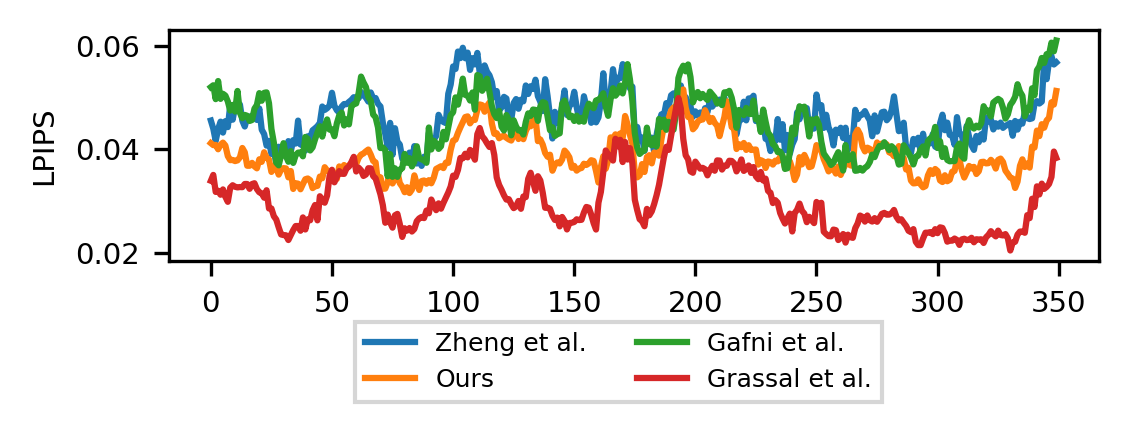}
    }
    \vfill
    \subfloat[]{
        \includegraphics[width=0.45\linewidth]{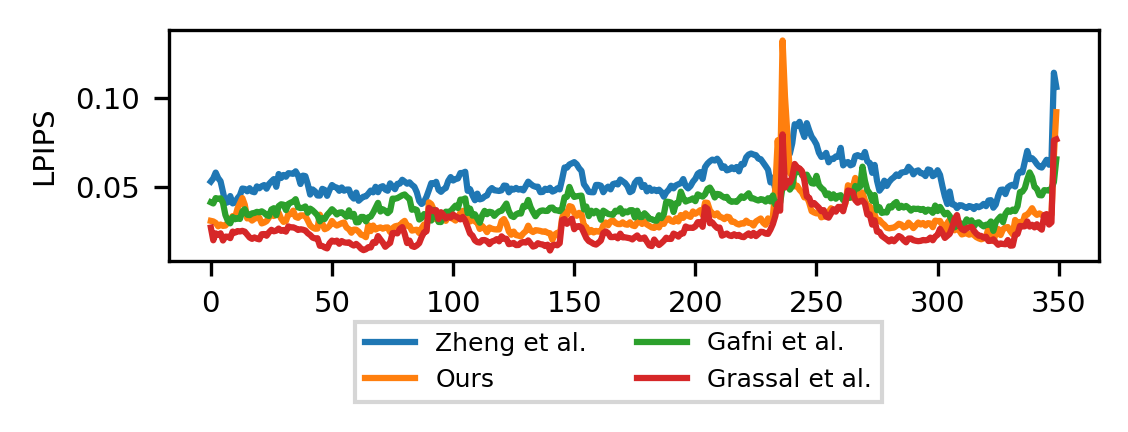}
    }
    \subfloat[]{
        \includegraphics[width=0.45\linewidth]{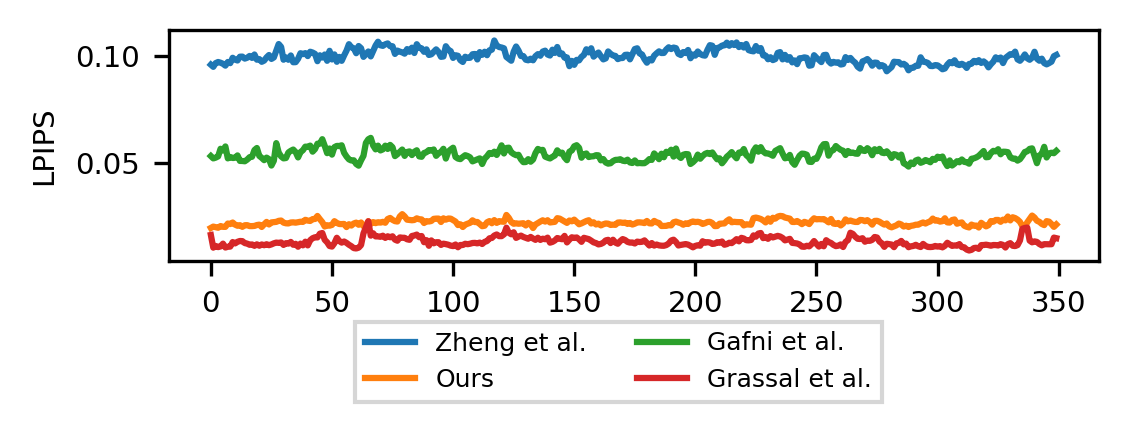}
    }
    \vfill
    \subfloat[]{
        \includegraphics[width=0.45\linewidth]{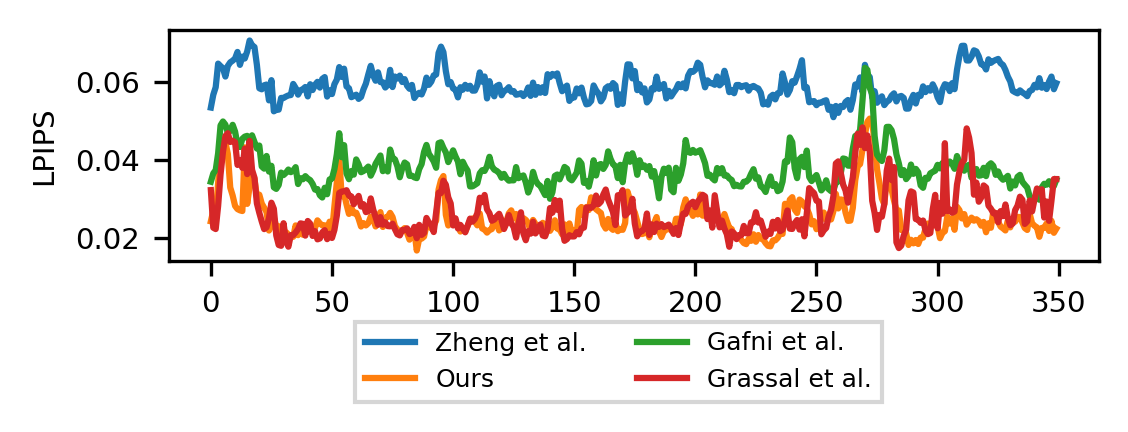}
    }
    \subfloat[]{
        \includegraphics[width=0.45\linewidth]{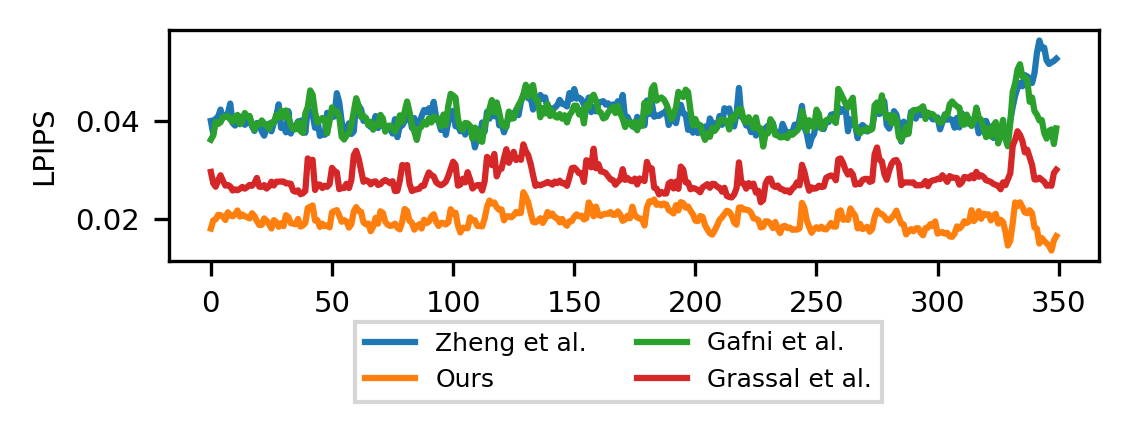}
    }
    \vfill
    \subfloat[]{
        \includegraphics[width=0.45\linewidth]{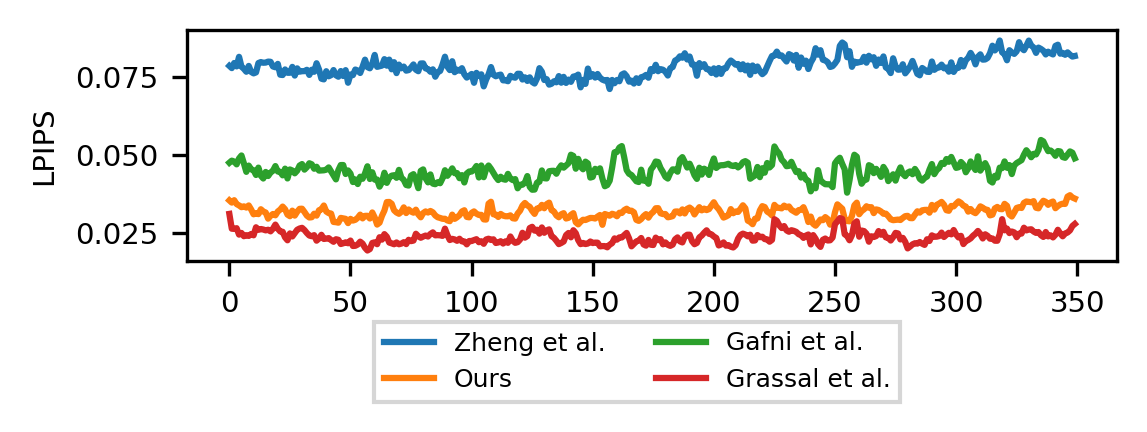}
    }
    \subfloat[]{
        \includegraphics[width=0.45\linewidth]{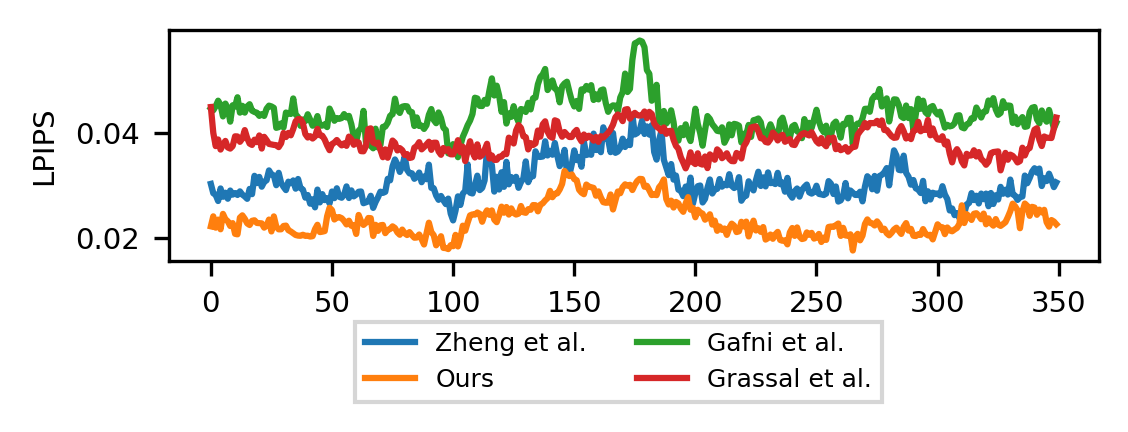}
    }
    \\
    \caption{Evaluation of the perceptual error for each of the volumetric avatars from our dataset on the test sequence. The alphanumerical order matches Figure \ref{fig:supp-teaser}. Our method achieves the lowest errors for color reconstruction and captures well even high-frequency details like freckles or wrinkles.}
    \label{fig:trcking_error}
\end{figure*}

\begin{figure*}[ht!]
    \centering
    \vspace{-0.5cm}
    \includegraphics[width=0.975\textwidth]{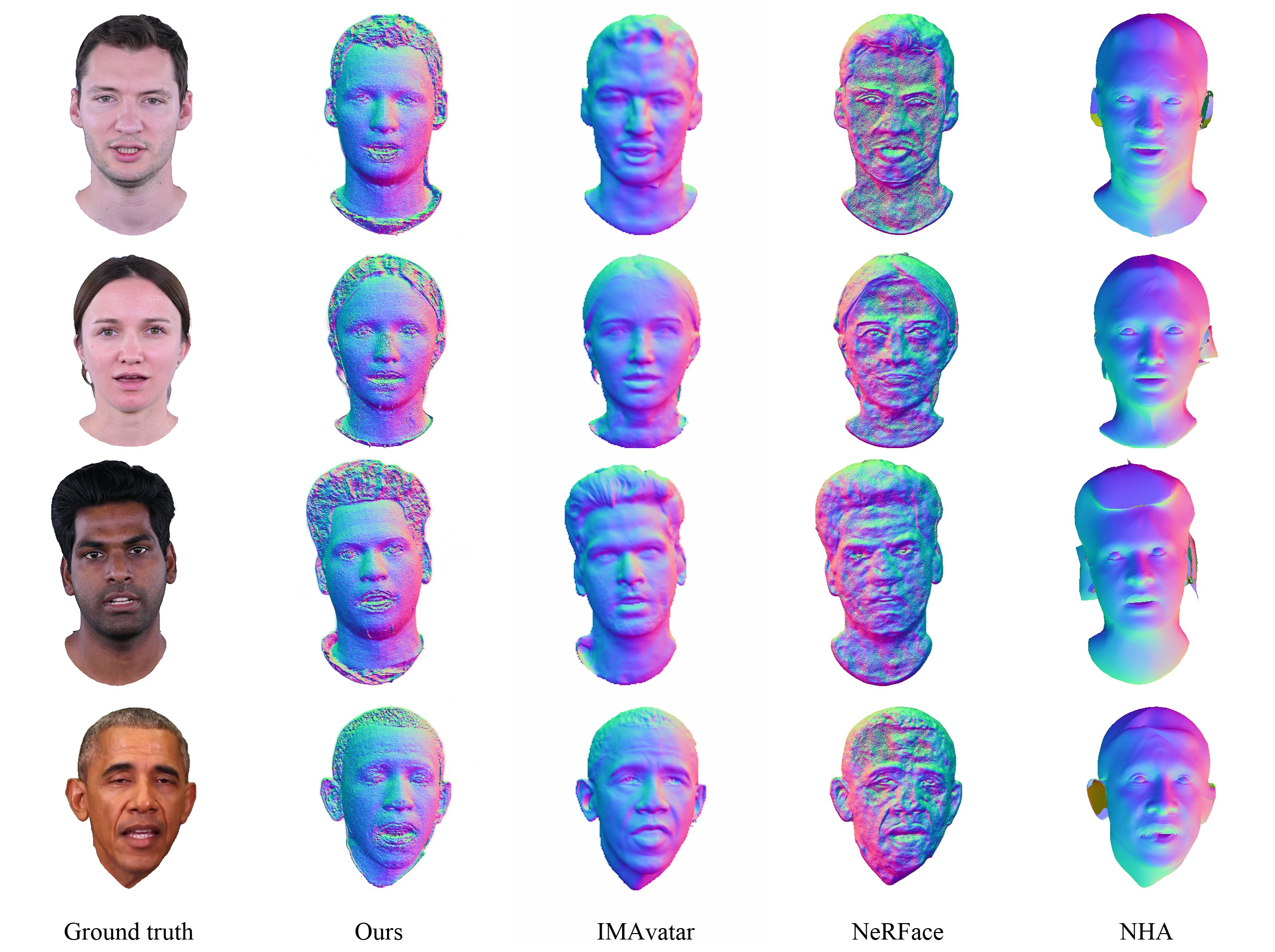}
    \caption{NeRF \cite{nerf} produces noisy normals maps, which can be seen in the results of Gafni et al. Our method uses additional geometric prior, which helps reduce the noise in specific areas, however, regions like hair are still problematic. The best results are achieved by IMAvatar \cite{imavatar}, which is based on a modified version of the IDR \cite{idr} approach. However, it takes a few days to achieve this result, while ours reaches similar quality, especially, in the face region, in about 10min.} 
    \label{fig:normals}
\end{figure*}

In \Cref{fig:normals}, we present estimated geometry as normal maps.
NeRF \cite{nerf} can recover geometry, but it contains noise.
Therefore, NeRFace~\cite{nerface}, which is based on NeRF, inherits the same problem.
In contrast, NHA \cite{nha} uses an explicit mesh representation based on the FLAME topology.
However, NHA produces deformed geometry, especially, for the ears.
Moreover, the optimized geometry is low-quality and misses many details, which are compensated by neural textures in the final image synthesis.
IMAvatar \cite{imavatar} is able to recover high-quality geometry.
The approach, based on IDR \cite{idr}, can take up to several days to optimize the dynamic occupancy field.
In contrast, our approach only needs a fraction of this time to optimize an avatar.
In the face region, the geometry quality is on par with IMAvatar. However, at the hair region where we do not have access to a geometric prior, the quality is similar to NeRFace.
In addition to the normal maps, we show the photo-metric error for a single frame in \Cref{fig:heatmaps} using an RGB-base $\ell_1$ metric; a perceptual evaluation for the entire sequences is shown in \Cref{fig:trcking_error}.

\section{Training Time Evaluation}

In \Cref{tab:training_time}, we present the average training times for each method.
Our method uses a local gaming PC equipped with a modern GPU Nvidia RTX 3090 and requires about \runtime min to reconstruct a volumetric avatar with high-frequency details.
For the baselines, we use their original configurations on a compute cluster.
Specifically, an Nvidia Quadro 6000 was used for the single GPU methods~\cite{imavatar,nerface}, and for NHA~\cite{nha}, three Nvidia A100 40GB GPUs were used.
While running on commodity hardware, our method is orders of magnitude faster than the others, making it more versatile and energy-saving.

\begin{table}[t!]
    \centering
    \begin{tabular}{llll}
        Method & Time & Units & GPUs \\
        \midrule
        IMAvatar \cite{imavatar}   &  $\sim4$ & day & 1 \\
        NeRFace \cite{nerface}   &  $\sim3$ & day & 1 \\
        NHA \cite{nha} &  $\sim13$ & hour & 3 \\
        Ours           &  $\sim$ \runtime & \textbf{minute} & 1 \\
        \bottomrule
    \end{tabular}
    \caption{
        Average training times for the avatar creation at a resolution of $512^2$, except IMAvatar which is using $256^2$. Note that the dataset generation for each of the methods is not taken into a consideration and only the avatar training time is measured.
    }
    \vspace{-0.55cm}
    \label{tab:training_time}
\end{table}


\section{Broader Impact}
\label{sec:impact}
\model synthesizes photo-realistic volumetric avatars from monocular RGB images and can extrapolate to novel views utilizing the 3DMM geometry prior.
Since \model does not require sophisticated capture setups, it can be applied to standard videos that can be captured with a webcam or a smartphone or downloaded from YouTube.
While our research focuses mainly on connecting people via teleconferencing, there is a risk of misuse.
Specifically, our method could be abused to produce so-called DeepFakes, which can be used for misinformation, cyber mobbing, identity theft, or other harmful criminal acts.
Unfortunately, we are not able to prevent the misuse of our technology.
However, conducting research openly and transparently could raise awareness of nefarious uses.
We will share our codebase to enable research on digital multi-media forensics, where synthesis methods are needed to produce a training corpus for forgery detection~\cite{roessler2019faceforensics++}.
All participants of our in-house recordings in the study have given written consent to the usage of their video material for this publication. YouTube videos were taken from public domains.

\end{document}